\documentclass[runningheads]{llncs}
\usepackage{graphicx}
\usepackage{tikz}
\usepackage{comment}
\usepackage{amsmath,amssymb} % define this before the line numbering.
\usepackage{color}

\usepackage[accsupp]{axessibility}  % Improves PDF readability for those with disabilities.

\usepackage[colorlinks=true]{hyperref}
\usepackage{cite}   % multiple cite in one
\usepackage{booktabs}  % row line in table
\usepackage{multirow}  % table multi row
\usepackage{pifont}    % c and x
\usepackage{makecell}
\usepackage[misc,geometry]{ifsym}  % letter for corresponding author
\newcommand{\etal}{\textit{et al.}} 

\newcommand{\cmark}{\text{\ding{51}}}
\newcommand{\xmark}{\text{\ding{55}}}

\begin{document}
% \linenumbers
\pagestyle{headings}
\mainmatter
\def\ECCVSubNumber{5048}  % Insert your submission number here

\title{CoMER: Modeling Coverage for Transformer-based Handwritten Mathematical Expression Recognition} % Replace with your title

\begin{comment}
\titlerunning{ECCV-22 submission ID \ECCVSubNumber} 
\authorrunning{ECCV-22 submission ID \ECCVSubNumber} 
\author{Anonymous ECCV submission}
\institute{Paper ID \ECCVSubNumber}
\end{comment}
%******************

%\begin{comment}
\titlerunning{CoMER: Modeling Coverage for Transformer-based HMER}
\author{Wenqi Zhao\inst{1} \and
Liangcai Gao\inst{1}(\Letter)}
\authorrunning{W. Zhao et al.}
\institute{Wangxuan Institute of Computer Technology, Peking University, Beijing, China \\
\email{wenqizhao@stu.pku.edu.cn, gaoliangcai@pku.edu.cn}}
%\end{comment}
%******************
\maketitle

\begin{abstract}
The Transformer-based encoder-decoder architecture has recently made significant advances in recognizing handwritten mathematical expressions. However, the transformer model still suffers from the lack of coverage problem, making its expression recognition rate (ExpRate) inferior to its RNN counterpart. Coverage information, which records the alignment information of the past steps, has proven effective in the RNN models. In this paper, we propose CoMER, a model that adopts the coverage information in the transformer decoder. Specifically, we propose a novel Attention Refinement Module (ARM) to refine the attention weights with past alignment information without hurting its parallelism. Furthermore, we take coverage information to the extreme by proposing self-coverage and cross-coverage, which utilize the past alignment information from the current and previous layers. Experiments show that CoMER improves the ExpRate by 0.61\%/2.09\%/1.59\% compared to the current state-of-the-art model, and reaches 59.33\%/59.81\%/62.97\% on the CROHME 2014/2016/2019 test sets.\footnote{Source code is available at \url{https://github.com/Green-Wood/CoMER}}

\end{abstract}

\section{Introduction}
Handwritten mathematical expression recognition (HMER) aims to generate the corresponding \LaTeX{} sequence from a handwritten mathematical expression image. The recognition of handwritten mathematical expressions has led to many downstream applications, such as online education, automatic scoring, and formula image searching. During the COVID-19 pandemic, an increasing number of education institutions chose to use online platforms for teaching and examing. The recognition rate of handwritten mathematical expressions is crucial to improving both learning efficiency and teaching quality in online education scenarios.

Handwritten mathematical expression recognition is an image-to-text task with more challenges than traditional text recognition. Besides various writing styles, we also need to model the relationships between symbols and contexts~\cite{anderson1967syntax}. In \LaTeX{}, for example, the model needs to generate ``\textasciicircum",  ``\textunderscore",  ``\{", and ``\}" to describe the position and hierarchical relationship between symbols in a two-dimensional image. Researchers use the encoder-decoder architecture widely in the HMER task~\cite{zhang2017watch, zhang2018multi, zhang2020treedecoder, wu2020handwritten, truong2020improvement, li2020improving, ding2021encoder, zhao2021handwritten} because of its feature extraction in the encoder part and language modeling in the decoder part.

Transformer~\cite{AshishVaswani2017AttentionIA}, a neural network architecture based solely on the attention mechanism, has gradually replaced RNN as the preferred model in natural language processing (NLP)~\cite{JacobDevlin2018BERTPO}. Through the self-attention mechanism in the transformer, tokens in the same sequence establish direct one-to-one connections. Such an architecture allows the transformer to better model long-term dependency~\cite{bengio1993problem} between tokens. Currently, Transformer is attracting more and more attention in the computer vision~\cite{DBLP:conf/iclr/DosovitskiyB0WZ21} and multimodal~\cite{DBLP:conf/aaai/LuoJSCWHLJ21, DBLP:conf/cvpr/PanYLM20, cornia2020meshed} community.

Although transformer has become the standard de-facto in NLP, its performance in the HMER task was unsatisfactory compared with its RNN counterparts~\cite{ding2021encoder, zhao2021handwritten}. We observe that the existing model using the transformer decoder still suffers from the lack of coverage problem~\cite{zhang2017watch, ZhaopengTu2016ModelingCF}. This problem manifests itself in two ways: over-parsing means that some parts of the image are unnecessarily pasred multiple times, while under-parsing means that some areas remain unparsed. RNN decoder uses coverage attention~\cite{zhang2017watch, zhang2018multi, zhang2020treedecoder, wu2020handwritten, truong2020improvement, li2020improving, ding2021encoder} to alleviate this problem. However, the current transformer decoder uses vanilla dot-product attention without the coverage mechanism, which is the key factor limiting its performance.

The computation of each step in the transformer is independent of each other, unlike RNN, where the computation of the current step depends on the previous step's state. While this nature improves the parallelism in the transformer, it makes it difficult to use the coverage mechanism from previous works directly in the transformer decoder. To address the above issues, we propose a novel model for exploiting \textbf{Co}verage information in the transfor\textbf{MER} decoder, named CoMER. Inspired by the coverage mechanism in RNN, we want the transformer to allocate more attention to regions that have not yet been parsed. Specifically, we propose a novel and general Attention Refinement Module (ARM) that dynamically refines the attention weights with past alignment information without hurting its parallelism. To fully use the past alignment information generated from different layers, we propose self-coverage and cross-coverage to utilize the past alignment information from the current and previous layer, respectively. We further show that CoMER performs better than vanilla transformer decoder and RNN decoder in the HMER task. The main contributions of our work are summarized as follows:

\begin{itemize}
	\item We propose a novel and general Attention Refinement Module (ARM) to refine the attention weight in the transformer decoder, which effectively alleviates the lack of coverage problem without hurting its parallelism.
	\item We propose self-coverage, cross-coverage, and fusion-coverage to fully use the past alignment information generated from different layers in the stack transformer decoder.
	\item Experiments show that CoMER outperforms existing state-of-the-art methods and achieves expression recognition rates (ExpRate)s of 59.33\%/ 59.81\%/ 62.97\% on the CROHME 2014~\cite{mouchere2014icfhr}/2016~\cite{mouchere2016icfhr2016}/2019~\cite{mahdavi2019icdar} datasets.
\end{itemize}

\section{Related Work}
\subsection{HMER Methods}
The traditional approach usually divides the HMER task into two subtasks: symbol recognition and structure analysis~\cite{chan2000mathematical}. Researchers represented the structural information of formulas through different predefined grammars, such as graph grammar~\cite{lavirotte1998mathematical}, context-free grammar~\cite{alvaro2014recognition}, and relational grammar~\cite{maclean2013new}. These methods require researchers to develop hand-designed grammar rules,  and their generalizability largely depends on the perfection of these grammar rules.

In recent years, encoder-decoder architectures have shown promising performance in various image-to-text tasks, such as scene text recognition~\cite{cheng2017focusing} and image captioning~\cite{xu2015show}. In~\cite{zhang2017watch}, a model called WAP was proposed to use encoder-decoder neural network for the first time to solve the HMER task and outperformed traditional grammar-based methods in the CROHME 2014 competition~\cite{mouchere2014icfhr}. The WAP model uses a convolution neural network (CNN) encoder, a gated recurrent unit (GRU) decoder, and coverage attention to form the encoder-decoder architecture. 

In terms of model architecture improvement, Zhang \etal~\cite{zhang2018multi} proposed DenseWAP, which uses a multi-scale DenseNet~\cite{GaoHuang2017DenselyCC} encoder to improve the ability to handle multi-scale symbols. Ding \etal~\cite{ding2021encoder} then borrows the architecture design of the transformer to improve the RNN-based model performance by multi-head attention and stacked decoder.

In terms of data augmentation, Li \etal~\cite{li2020improving} proposed scale augmentation that scales the image randomly while maintaining the aspect ratio, which improves the generalization ability over multi-scale images. PAL-v2~\cite{wu2020handwritten} then uses printed mathematical expressions as additional data to help train the model.

In terms of training strategies, Truong \etal~\cite{truong2020improvement} proposed WS-WAP by introducing weakly supervised information about the presence or absence of symbols to the encoder. Besides,  BTTR~\cite{zhao2021handwritten} was proposed to first use the transformer decoder for solving HMER task, and perform bidirectional language modeling with a single decoder.

\subsection{Coverage Mechanism}
The coverage mechanism was first proposed~\cite{ZhaopengTu2016ModelingCF} to solve the over-translation and under-translation problems in the machine translation task. 

All of the previous works in HMER~\cite{ZhaopengTu2016ModelingCF, zhang2017watch, zhang2018multi, zhang2020treedecoder, wu2020handwritten, truong2020improvement, li2020improving, ding2021encoder} used the coverage attention in RNN, where a coverage vector is introduced to indicate whether an image feature vector has been parsed or not, leading the model to put more attention on the unparsed regions. It is a step-wise refinement, where the decoder needs to collect past alignment information for each step. For the RNN model, the decoder can naturally accumulate the attention weights in each step, but it is difficult for the transformer decoder which performs parallel decoding. 

There is a work~\cite{rosendahl-etal-2021-recurrent} that tried to introduce the coverage machenism in transformer decoder. They directly used the coverage mechanism in RNN to the transformer, which greatly hurts its parallelism and training efficiency. Our CoMER model, on the other hand, utilizes the coverage information as an attention refinement term in the transformer decoder without hurting its parallel decoding nature.

\begin{figure}[htbp]
	\centering
	\includegraphics[width=0.6\textwidth]{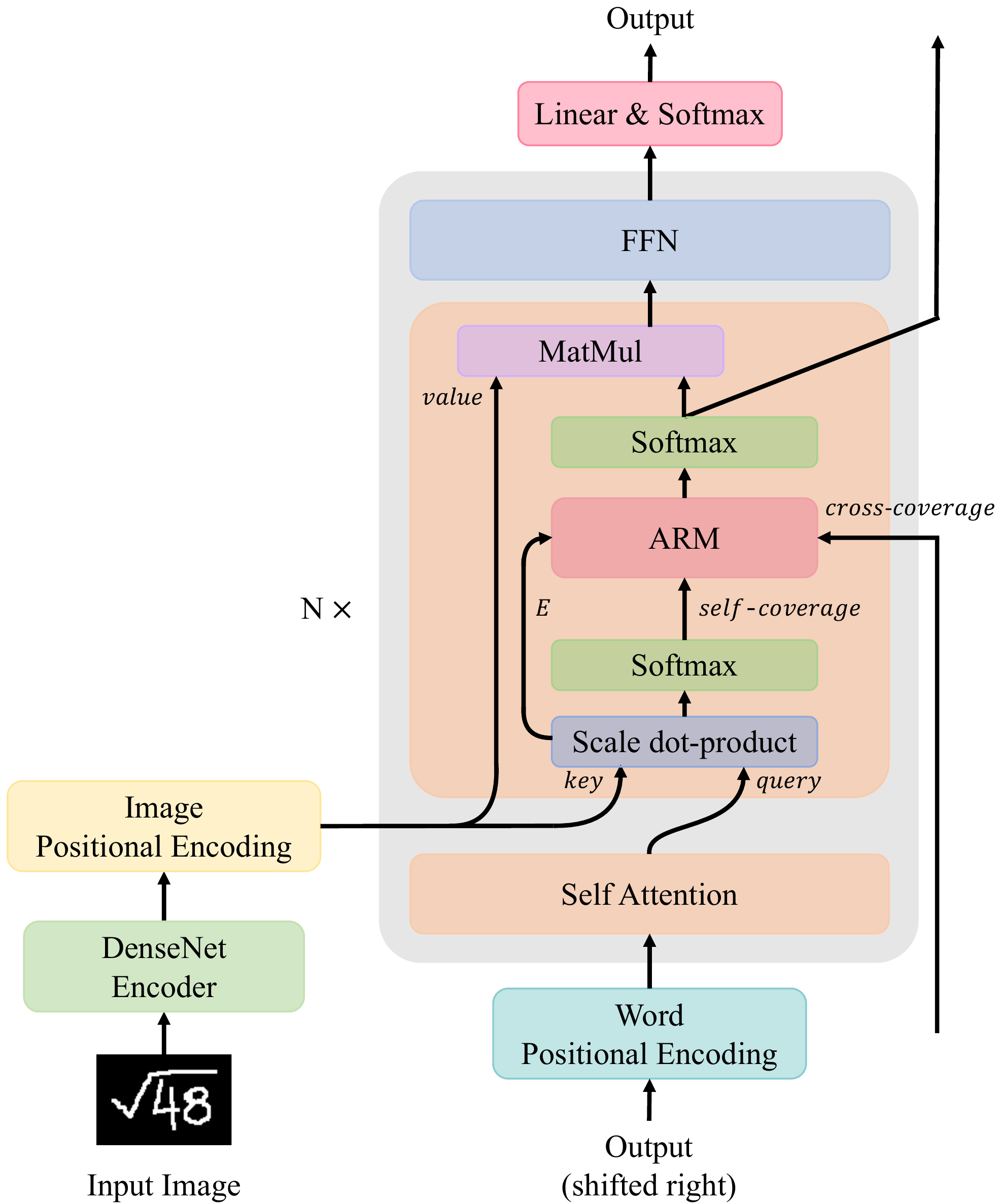}
	\caption{The overview architecture of our proposed CoMER model. The attention weights generated by $key$ and $query$ are fed into a novel Attention Refinement Module (ARM). ARM utilizes the past alignment information generated from different layers through self-coverage and cross-coverage.}
	\label{fig:CoMER}
\end{figure}

\section{Methodology}
In this section, we will first review the coverage attention in RNN and multi-head attention, then describe the architecture design of CoMER in detail. As illustrated in Fig.~\ref{fig:CoMER}, the model consists of four main modules: 1) CNN Encoder, which extracts features from 2D formula images. 2) Positional Encoding that addresses position information for the transformer decoder. 3) Attention Refinement Module (ARM) is used to refine the attention weights with the past alignment information. 4) Self-coverage and cross-coverage utilize the past alignment information from the current and previous layers.

\subsection{Background}

\subsubsection{Coverage Attention in RNN}
Coverage attention has been widely used in RNN-based HMER models~\cite{ZhaopengTu2016ModelingCF, zhang2017watch, zhang2018multi, zhang2020treedecoder, wu2020handwritten, truong2020improvement, li2020improving, ding2021encoder}. The coverage vector provides information to the attention model about whether a region has been parsed or not. Let encoder produces flatten output image feature $\mathbf{X}_f \in \mathbb{R}^{L \times d_\text{model}}$ with the sequence length $L = h_o \times w_o$. At each step $t$, previous attention weights $\mathbf a_k$ are accumulated as vector $\mathbf c_t$, and then transformed into coverage matrix $\mathbf F_t$.
\begin{equation}\label{eq:rnn-cumsum}
	\mathbf c_t = \sum_{k=1}^	{t-1} \mathbf{a}_k \in \mathbb{R}^{L}
\end{equation}
\begin{equation}\label{eq:rnn-cov}
	\mathbf F_t = \operatorname{cov}(\mathbf c_t) \in \mathbb{R}^{L \times d_\text{attn}}
\end{equation}
here $\operatorname{cov}(\cdot)$ denotes a composite function of an $11 \times 11$ convolution layer and a linear layer. 

In attention mechanism, we calculate a similarity score $e_{t, i}$ for every image feature at index $i \in [0,L)$. By taking advantage of broadcast operations in modern deep learning frameworks, such as PyTorch~\cite{AdamPaszke2019PyTorchAI}, we can compute similarity vector $\mathbf e_t$ in parallel by broadcasting RNN hidden state $\mathbf{h}_t \in \mathbb{R}^{d_\text{model}}$ to $\mathbf{H}_t \in \mathbb{R}^{L \times d_\text{model}}$. The attention weights $\mathbf a_t$ of current step $t$ is obtained as follow:
\begin{equation}\label{eq:rnn-attention}
	\mathbf{e}_t = \tanh \left(\mathbf{H}_{t}\mathbf{W}_{h} + \mathbf{X}_{f}\mathbf{W}_{x} + \mathbf{F}_{t}\right)\mathbf{v}_{a}
\end{equation}
\begin{equation}
	\mathbf a_t = \operatorname{softmax}(\mathbf e_t) \in \mathbb{R}^{L}
\end{equation}
where $\mathbf{W}_{h} \in \mathbb{R}^{d_\text{model} \times d_\text{attn}}$, $\mathbf{W}_{x} \in \mathbb{R}^{d_\text{model} \times d_\text{attn}}$ are trainable parameters matrices, and $\mathbf v_a \in \mathbb{R}^{d_\text{attn}}$ is a trainable parameter vector.

\subsubsection{Multi-Head Attention} 
Multi-head attention is the most critical component of the transformer models~\cite{AshishVaswani2017AttentionIA}. 
With the model dimension size $d_\text{model}$, query sequence length $T$, and key sequence length $L$, we split the multi-head attention calculation for head $\mathbf{Head}_i$ into four parts: 1) project the query $\mathbf{Q}$, key $\mathbf{K}$, and value $\mathbf{V}$ into a subspace; 2) calculate the scaled dot-product $\mathbf{E}_i \in \mathbb{R}^{T \times L}$; 3) compute the attention weights $\mathbf{A}_i \in \mathbb{R}^{T \times L}$ by the softmax function; 4) obtain head $\mathbf{Head}_i$ by multiplying the attention weights $\mathbf{A}_i$ and value $\mathbf{V}_i$.
\begin{equation}
	\mathbf{Q}_i, \mathbf{K}_i, \mathbf{V}_i = \mathbf Q \mathbf W_{i}^{Q}, \mathbf K \mathbf W_{i}^{K},\mathbf V \mathbf W_{i}^{V}
\end{equation}
\begin{equation}\label{eq:scale_dot_prod}
	\mathbf{E}_i = \frac{\mathbf Q_i \mathbf K^\intercal_i}{\sqrt{d_{k}}} \in \mathbb{R}^{T \times L}
\end{equation}
\begin{equation}\label{eq:softmax_attn}
	\mathbf{A}_i = \operatorname{softmax}(\mathbf{E}_i) \in \mathbb{R}^{T \times L}
\end{equation}
\begin{equation}
	\mathbf{Head}_i = \mathbf{A}_i\mathbf{V}_i \in \mathbb{R}^{T \times d_v}
\end{equation}
where 
$
\mathbf W_{i}^{Q} \in \mathbb{R}^{d_{\text {model }} \times d_{k}}, 
\mathbf W_{i}^{K} \in \mathbb{R}^{d_{\text {model }} \times d_{k}}, 
\mathbf W_{i}^{V} \in \mathbb{R}^{d_{\text {model }} \times d_{v}}
$
denote the trainable projection parameter matrices. Then all $h$ heads are concatenated and projected with a trainable projection matrix $\mathbf W^{O} \in \mathbb{R}^{h d_{v} \times d_{\text {model }}}$ to obtain the final output:
\begin{equation}
	\operatorname{MultiHead}(\mathbf Q, \mathbf K, \mathbf V)=\left[\mathbf{Head}_1; \ldots; \mathbf{Head}_{h}\right] \mathbf W^{O}
\end{equation}

We follow this setting in CoMER and use the Attention Refinement Module (ARM) in Sect.~\ref{sec:ARM} to refine the scale dot-product matrix $\mathbf{E}_i$ in Eq.~\eqref{eq:scale_dot_prod}.

\subsection{CNN Encoder}
In the encoder part, we use DenseNet~\cite{GaoHuang2017DenselyCC} to extract features in the 2D formula image, following the same setting with BTTR~\cite{zhao2021handwritten}. The core idea of DenseNet is to enhance the information flow between layers through concatenation operation in the feature dimension. Specifically, in the DenseNet block $b$, the output feature of $l^{th}$ layer can be computed by the output features $\mathbf{X}_{0}, \mathbf{X}_{1}, \ldots, \mathbf{X}_{l-1} \in \mathbb{R}^{h_b \times w_b \times d_b}$ from the previous $0^{th}$ to $(l-1)^{th}$ layers:
\begin{equation}
	\mathbf{X}_{\ell}=H_{\ell}\left(\left[\mathbf{X}_{0}; \mathbf{X}_{1}; \ldots; \mathbf{X}_{\ell-1}\right]\right) \in \mathbb{R}^{h_b \times w_b \times d_b}
\end{equation}
where $\left[\mathbf{X}_{0}; \mathbf{X}_{1}; \ldots; \mathbf{X}_{\ell-1}\right] \in \mathbb{R}^{h_b \times w_b \times (ld_b)}$ denotes the concatenation operation in the feature dimension, $d_b$ denotes the feature dimension size of DenseNet block, and $H_{\ell}(\cdot)$ function is implemented by: a batch normalization~\cite{SergeyIoffe2015BatchNA} layer, a ReLU~\cite{XavierGlorot2011DeepSR} activation function, and a $3 \times 3$ convolution layer. 

In order to align DenseNet output feature with the model dimension size $d_\text{model}$, we add a $1 \times 1$ convolution layer at the end of the encoder to obtain the output image feature $\mathbf{X}_o \in \mathbb{R}^{h_o \times w_o \times d_\text{model}}$.

\subsection{Positional Encoding}
Unlike the RNN decoders, which inherently consider the order of word tokens, the additional position information is necessary for the transformer decoder due to its permutation-invariant property. In CoMER, we are consistent with BTTR~\cite{zhao2021handwritten}, employing both image positional encoding and word positional encoding.

For word positional encoding, we use the 1D positional encoding introduced in the vanilla transformer~\cite{AshishVaswani2017AttentionIA}. Given encoding dimension size $d$, position $p$, and the index $i$ of feature dimension, the word positional encoding vector $\mathbf{p}^{\mathbf{W}}_{p, d} \in \mathbb{R}^{d}$ can be represented as:
\begin{equation} \label{eq:word_pos_enc}
\begin{split}
	\mathbf{p}^{\mathbf{W}}_{p, d}[2 i] &= \sin (p / 10000^{2 i / d}) \\
	\mathbf{p}^{\mathbf{W}}_{p, d}[2 i + 1] &= \cos (p / 10000^{2 i / d})
\end{split}
\end{equation}

For image positional encoding, a 2D normalized positional encoding is used, which is the same as~\cite{zhao2021handwritten, NicolasCarion2020EndtoEndOD}. Since it is not the absolute position but the relative position that matters, the position coordinates should be normalized first. Given a 2D coordinates tuple $(x, y)$ and the encoding dimension size $d$, the image positional encoding $\mathbf{p}^{\mathbf{I}}_{x,y,d} \in \mathbb{R}^{d}$ is computed by the concatenation of 1D positional encoding~\eqref{eq:word_pos_enc} of two dimensions.
\begin{equation}
	\bar{x} = \frac{x}{h_o}, \quad \bar{y} = \frac{y}{w_o}
\end{equation}
\begin{equation}
	\mathbf{p}^{\mathbf{I}}_{x,y,d} = [\mathbf{p}^{\mathbf{W}}_{\bar{x}, d/2}; \mathbf{p}^{\mathbf{W}}_{\bar{y}, d/2}]
\end{equation}
where $h_o$ and $w_o$ denote the shape of output image feature $\mathbf{X}_o \in \mathbb{R}^{h_o \times w_o \times d_\text{model}}$.

\subsection{Attention Refinement Module} \label{sec:ARM}
Although coverage attention has been widely used in the RNN decoder, it is difficult to use it directly in the transformer decoder due to transformer's parallel decoding. The inability to model the coverage information directly in the transformer leads to its unsatisfactory performance in the HMER task. We will first introduce the difficulty of using coverage information in the transformer in this subsection, then propose a novel Attention Refinement Module (ARM) to solve it. 

A naive solution is to use the multi-head attention weight $\mathbf A$ in Eq.~\eqref{eq:softmax_attn}, accumulate it as $\mathbf C$, and then transform into a coverage matrix $\mathbf F$ using $\operatorname{cov}(\cdot)$ function in Eq.~\eqref{eq:rnn-cov}. However, this naive solution is unacceptable considering space complexity. Assume that multi-head attention weight $\mathbf A \in \mathbb{R}^{T \times L \times h}$, then $\operatorname{cov}(\cdot)$ function will be applied at every time step and every image feature location, which will produce coverage matrix $\mathbf F \in \mathbb{R}^{T \times L \times h \times d_\text{attn}}$ with space complexity $O(TLhd)$.

We can see that the bottleneck comes from the $\tanh(\cdot)$ function in Eq.~\eqref{eq:rnn-attention} where the coverage matrix needs to be summed with other feature vectors first, then multiplied by vector $\mathbf v_a \in \mathbb{R}^{d_\text{attn}}$. If we can multiply the coverage matrix with $\mathbf v_a$ first and then add the result of LuongAttention~\cite{MinhThangLuong2015EffectiveAT}, the space complexity will be greatly reduced to $O(TLh)$. So we modify Eq.~\eqref{eq:rnn-attention} as follows:
\begin{equation} \label{eq:attention_refinement}
\begin{split}
	\mathbf{e}_t^\prime &= \tanh \left(\mathbf{H}_{t}\mathbf{W}_{h} + \mathbf{X}_{f}\mathbf{W}_{x}\right)\mathbf{v}_{a} + \mathbf{F}_{t}\mathbf{v}_{a} \\
	&= \underbrace{\tanh \left(\mathbf{H}_{t}\mathbf{W}_{h} + \mathbf{X}_{f}\mathbf{W}_{x}\right)\mathbf{v}_{a}}_{\text{attention}} + \underbrace{\mathbf r_t }_{\text{refinement}}
\end{split}
\end{equation}
where similarity vector $\mathbf e_t^\prime$ can be divided into an attention term and a refinement $\mathbf r_t \in \mathbb{R}^{L}$ term. Notice that given accumulated vector $\mathbf c_t$ in Eq.~\eqref{eq:rnn-cumsum}, refinement term $\mathbf r_t$ could be directly produced by a coverage modeling function, avoiding intermediate representation with dimension $d_\text{attn}$. We name the process in Eq.~\eqref{eq:attention_refinement} as the \textbf{Attention Refinement Framework}.

\begin{figure}[htbp]
	\centering
	\includegraphics[width=0.6\textwidth]{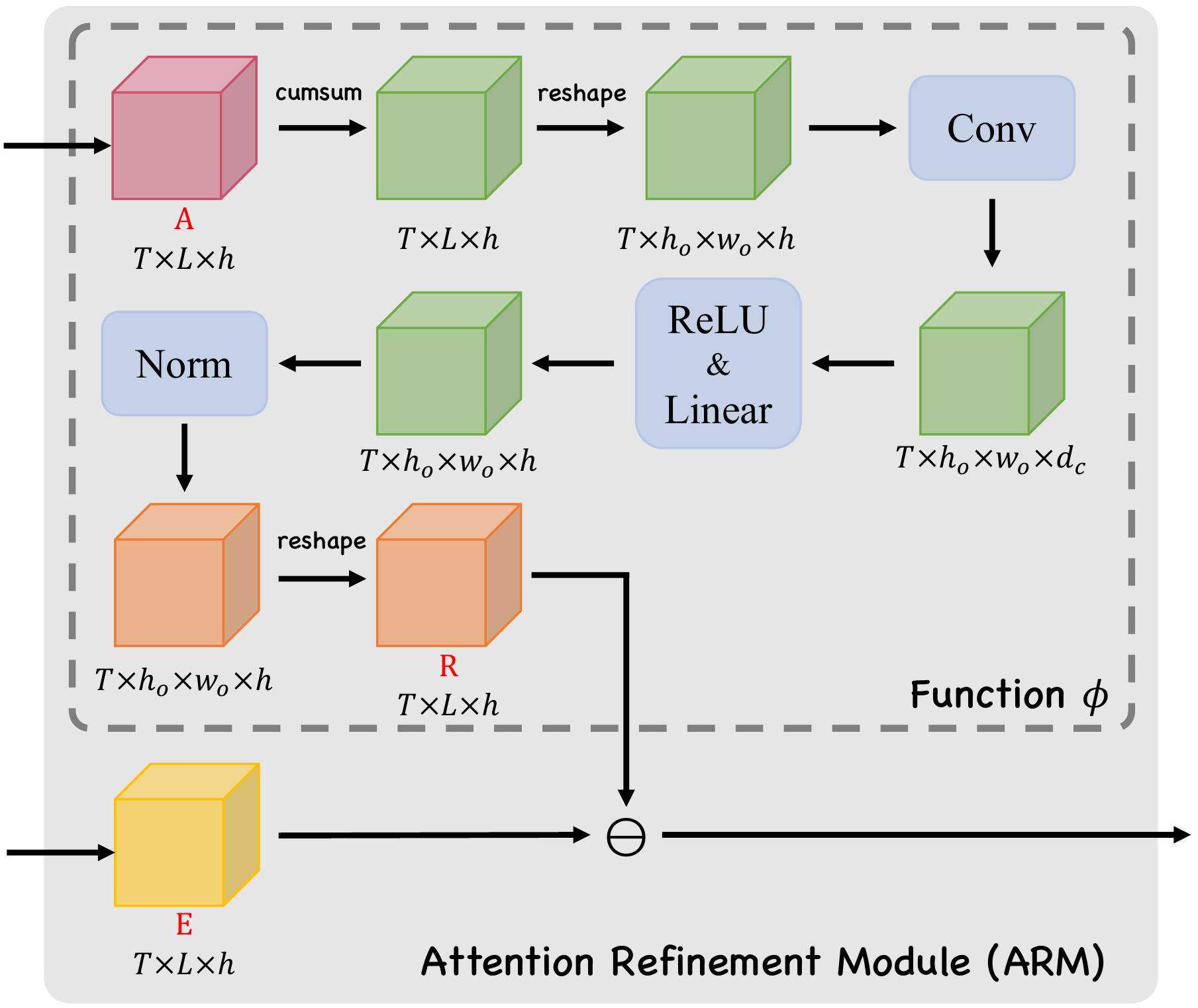}
	\caption{The overview of  Attention Refinement Module (ARM). Given the generic attention weights $\mathbf A$, we first calculate the refinement term $\mathbf R$ using function $\phi(\cdot)$ . Then, we refine the attention term $\mathbf E$ by subtracting the refinement term $\mathbf R$.}
	\label{fig:ARM}
\end{figure}

To use this framework in the transformer, we propose an \textbf{Attention Refinement Module (ARM)} shown in Fig.~\ref{fig:ARM}. The scale dot-product matrix $\mathbf E \in \mathbb{R}^{T \times L \times h}$ in Eq.~\eqref{eq:scale_dot_prod} can be used as the attention term, and the refinement term matrix $\mathbf R$ needs to be calculated from the attention weights $\mathbf A$. Note that we use a generic attention weights $\mathbf A$ here to provide past alignment information, and the specific choice of it will be discussed in Sect.~\ref{sec:coverage}. 

We define a function $\phi: \mathbb{R}^{T \times L \times h} \mapsto \mathbb{R}^{T \times L \times h}$ that takes the attention weights $\mathbf A \in \mathbb{R}^{T \times L \times h}$ as input and outputs the refinement matrix $\mathbf R \in \mathbb{R}^{T \times L \times h}$. With kernel size $k_c$, intermediate dimension $d_c \ll h \times d_\text{attn}$, and the output image feature shape $L = h_o \times w_o$, function $\phi(\cdot)$ is defined as:
\begin{equation} \label{eq:theta_func}
	\mathbf R = \phi(\mathbf A) = \operatorname{norm} \left( \operatorname{max} \left(0, \mathbf K * \widetilde{\mathbf C} + \mathbf b_c\right) \mathbf W_c \right)
\end{equation}
\begin{equation}
	\widetilde{\mathbf C} = \operatorname{reshape}(\mathbf C) \in \mathbb{R}^{T \times h_o \times w_o \times h}
\end{equation}
\begin{equation} \label{eq:mhattn_cumsum}
	\mathbf c_t = \operatorname{} \sum_{k=1}^	{t-1} \mathbf{a}_k \in \mathbb{R}^{L \times h}
\end{equation}
where $\mathbf a_t$ is the attention weights at step $t \in [0, T)$, $\mathbf K \in \mathbb{R}^{k_c \times k_c \times h \times d_c}$ denotes a convolution kernel, $*$ denotes convolution operation over the reshaped accumulated matrix $\widetilde{\mathbf C} \in \mathbb{R}^{T \times h_o \times w_o \times h}$, $\mathbf b_c \in \mathbb{R}^{d_c}$ is a bias term, and $\mathbf W_c \in \mathbb{R}^{d_c \times h}$ is a linear projection matrix. Note that Eq.~\eqref{eq:mhattn_cumsum} can be efficiently computed by $\operatorname{cumsum}(\cdot)$ function in modern deep learning frameworks~\cite{AdamPaszke2019PyTorchAI}.

We consider that function $\phi$ can extract local coverage features to detect the edge of parsed regions and identify the incoming unparsed regions. Finally, we refine the attention term $\mathbf E$ by subtracting the refinement term $\mathbf R$. 
\begin{equation}\label{eq:arm}
\begin{split}
	\operatorname{ARM}(\mathbf E, \mathbf A) &= \mathbf E - \mathbf R \\
	&= \mathbf E - \phi(\mathbf A)
\end{split}
\end{equation}

\subsection{Coverage}\label{sec:coverage}
In this section, we will discuss the specific choice of the generic attention weights $\mathbf A$ in Eq.~\eqref{eq:theta_func} We propose self-coverage and cross-coverage to utilize alignment information from different stages, introducing diverse past alignment information to the model.

\subsubsection{Self-coverage} Self-coverage refers to using the alignment information generated by the current layer as input to the Attention Refinement Module. For the current layer $j$, we first calculate the attention weights $\mathbf A^{(j)}$, and refine itself.
\begin{equation}
	\mathbf{A}^{(j)} = \operatorname{softmax}(\mathbf{E}^{(j)}) \in \mathbb{R}^{T \times L \times h}
\end{equation}
\begin{equation}
	\mathbf{\hat{E}}^{(j)} = \operatorname{ARM} ( \mathbf E^{(j)}, \mathbf A^{(j)} )
\end{equation}
\begin{equation}
	\mathbf{\hat A}^{(j)} = \operatorname{softmax}(\mathbf{\hat E}^{(j)})
\end{equation}
where $\mathbf{\hat{E}}^{(j)}$ denotes the refined scale dot-product, and $\mathbf{\hat{A}}^{(j)}$ denotes the refined attention weights at layer $j$.

\subsubsection{Cross-coverage}
We propose a novel cross-coverage by exploiting the nature of the stacked decoder in the transformer. Cross-coverage uses the alignment information from the previous layer as input to the ARM of the current layer. For the current layer $j$, we use the refined attention weights $\mathbf{\hat A} ^{(j-1)}$ from the previous $(j - 1)$ layer and refine the attention term of the current layer.
\begin{equation}
	\mathbf{\hat{E}}^{(j)} = \operatorname{ARM} (\mathbf E^{(j)}, \mathbf{\hat A}^{(j-1)} )
\end{equation}
\begin{equation}
	\mathbf{\hat A}^{(j)} = \operatorname{softmax}(\mathbf{\hat E}^{(j)})
\end{equation}
Notice that $\mathbf{\hat A} ^{(j-1)} = \mathbf{A} ^{(j-1)}$ holds if the previous layer do not use the ARM.

\subsubsection{Fusion-coverage}
Combining the self-coverage and cross-coverage, we propose a novel fusion-coverage method to fully use the past alignment information generated from different layers.
\begin{equation}
	\mathbf{\hat{E}}^{(j)} = \operatorname{ARM} ( \mathbf E^{(j)}, [\mathbf A^{(j)}; \mathbf{\hat A}^{(j-1)} ])
\end{equation}
\begin{equation}
	\mathbf{\hat A}^{(j)} = \operatorname{softmax}(\mathbf{\hat E}^{(j)})
\end{equation}
where $[\mathbf A^{(j)}; \mathbf{\hat A}^{(j-1)}] \in \mathbb{R}^{T \times L \times 2h}$ denotes the concatenation of attention weights from the current layer and refined attention weights from the previous layer.

\section{Experiments}
\subsection{Implementation Details}
%To make a fair comparison with the baseline model BTTR~\cite{zhao2021handwritten}, which uses a vanilla transformer decoder, our CoMER follows the same setting with it.

In the encoder part, we employ the same DenseNet to extract features from the formula image. Three densenet blocks are used in the encoder, each containing $D = 16$ bottleneck layers. A transition layer is inserted between every two densenet blocks to reduce the spatial and channel size of the feature map by $\theta = 0.5$. The growth rate is set to $k = 24$, and the dropout rate is set to 0.2.

In the decoder part, for hyperparameters in the transformer decoder, we set the model dimension to $d_\text{model} = 256$, the number of heads to $h = 8$, and the feed-forward layer dimension size to $d_{ff} = 1024$. We use three stacked decoder layers and a 0.3 dropout rate. For the Attention Refinement Module that we proposed, we set the kernel size to $k_c = 5$, the intermediate dimension to $d_c = 32$. The normalization method we adopt in ARM is batch-normalization~\cite{SergeyIoffe2015BatchNA}. We use ARM starting with the second layer and share the same ARM between layers.

We use the same bidirectional training strategy as BTTR~\cite{zhao2021handwritten} to train CoMER with PyTorch~\cite{AdamPaszke2019PyTorchAI} framework. We use SGD with a weight decay of $10^{-4}$ and momentum of $0.9$. The learning rate is $0.08$. We augment input images using scale augmentation~\cite{li2020improving} with uniformly sampled scaling factor $s \in [0.7, 1.4]$. All experiments are conducted on four NVIDIA 2080Ti GPUs with $4 \times 11$ GB memory.

In the inference phase, instead of the beam search, we perform the approximate joint search~\cite{LemaoLiu2016AgreementOT} that has been used in BTTR~\cite{zhao2021handwritten}.

\subsection{Datasets and Metrics}
We use the Competition on Recognition of Online Handwritten Mathematical Expressions (CROHME) datasets~\cite{mouchere2014icfhr, mouchere2016icfhr2016, mahdavi2019icdar} to conduct our experiments, which is currently the largest open dataset for HMER task. The training set contains a total of 8836 training samples, while the CROHME 2014/2016/2019 test set contains 986/1147/1199 test samples. The CROHME 2014 test set~\cite{mouchere2014icfhr} is used as the validation set to select the best-performing model during the training process.

We use the evaluation tool officially provided by the CROHME 2019~\cite{mahdavi2019icdar} organizers to convert the predicted \LaTeX{} sequences into symLG format. Then, metrics are reported by utilizing the LgEval library~\cite{RichardZanibbi2013EvaluatingSP}. We choose ``ExpRate", ``$\leq 1$ error", ``$\leq 2$ error", and ``$\leq 3$ error" metrics to measure the performance of our proposed model. These metrics represent the expression recognition rate when we tolerate 0 to 3 symbol or structural errors.

\subsection{Ablation Study}
To verify the effectiveness of our proposed method, we performed ablation experiments on the CROHME 2014 test set~\cite{mouchere2014icfhr}. In Table~\ref{tb:ablation}, the ``Scale-aug" column indicates whether to adopt scale augmentation~\cite{li2020improving} for data augmentation. The ``Self-cov" column indicates whether self-coverage is used. The ``Cross-cov" column suggests the use of cross-coverage. 

First, since the original BTTR~\cite{zhao2021handwritten} did not use any data augmentation methods, we re-implement the BTTR model and achieve an ExpRate of 53.45\%, which is similar to the original result in~\cite{zhao2021handwritten}. To compare BTTR as a baseline with our proposed CoMER, we also use scale augmentation to train BTTR and obtained an ExpRate of 55.17\% for ``BTTR (baseline)".

The performance of CoMER using ARM and coverage mechanism has been significantly improved compared to BTTR. Comparing the last four rows in Table~\ref{tb:ablation}, we can observe that:
\begin{enumerate}
	\item When CoMER uses self-coverage to refine the attention weights, the performance is improved by 2.34\% compared to ``BTTR (baseline)". Experiment results validate the feasibility and effectiveness of using past alignment information in the transformer decoder.
	\item Compared to self-coverage, using cross-coverage in CoMER can bring more performance gains owing to the more accurate alignment information from the previous layer.
	\item ``CoMER (Fusion)" obtains the best results by combining self-coverage and cross-coverage, outperforming the baseline model by 4.16\%. This experiment results suggest that diverse alignment information generated from different layers helps ARM refine the current attention weights more accurately.
\end{enumerate}

\begin{table}[htbp]
\renewcommand{\arraystretch}{1.1}
\setlength{\tabcolsep}{4pt}
\caption{Ablation study on the CROHME 2014 test set (in \%). $\dagger$ denotes original reported results of BTTR~\cite{zhao2021handwritten}}
\begin{center}
\begin{tabular}{c|ccc|c}
\toprule 
Model & Scale-aug~\cite{li2020improving} & Self-cov & Cross-cov & ExpRate \\
\midrule
BTTR$^{\dagger}$~\cite{zhao2021handwritten} & \xmark & \xmark & \xmark & 53.96 \\
\midrule
BTTR & \xmark & \xmark & \xmark & 53.45 \\
BTTR (baseline) & \cmark & \xmark & \xmark & 55.17 (+0.00) \\
CoMER (Self) & \cmark & \cmark & \xmark & 57.51 (+2.34) \\
CoMER (Cross) & \cmark & \xmark & \cmark & 58.11 (+2.94) \\
CoMER (Fusion) & \cmark & \cmark & \cmark & \textbf{59.33} (+4.16) \\
\bottomrule
\end{tabular}
\label{tb:ablation}
\end{center}
\end{table}

\subsection{Comparison with State-of-the-art Approaches}
We compare the proposed CoMER with the previous state-of-the-art methods, as shown in Table~\ref{tb:sota}. For the RNN-based models, we choose DenseWAP~\cite{zhang2018multi}, DenseWAP-TD~\cite{zhang2020treedecoder}, WS-WAP~\cite{truong2020improvement}, Li \etal~\cite{li2020improving}, and Ding \etal~\cite{ding2021encoder} for comparison. For transformer-based models, we compare with BTTR~\cite{zhao2021handwritten} that uses a vanilla transformer decoder. Note that the approaches proposed by Li \etal~\cite{li2020improving} and Ding \etal~\cite{ding2021encoder} also use the scale augmentation~\cite{li2020improving}.

Compared with the RNN-based models that use coverage attention, CoMER outperforms the previous state-of-the-art model proposed by Ding \etal~\cite{ding2021encoder} on each CROHME test set. In the ExpRate metric, CoMER improves by an average of 1.43\% compared to the previous best-performing RNN-based model. 

Compared with the transformer-based model, our proposed CoMER equipped with ARM and fusion-coverage significantly improves the performance. Specifically, CoMER outperforms ``BTTR (baseline)" in all metrics and averages 3.6\% ahead of ``BTTR (baseline)" in ExpRate.

\begin{table}[htbp]
\renewcommand{\arraystretch}{0.9}
\setlength{\tabcolsep}{4pt}
\caption{Performance comparison with previous state-of-the-art approaches on the CROHME 2014/2016/2019 test sets (in \%).}
\begin{center}
\begin{tabular}{c|c|cccc}
\toprule 
Dataset & Model & ExpRate & $\leq 1$ error & $\leq 2$ error & $\leq 3$ error \\
\midrule

\multirow{8}{*}{CROHME 14}
& DenseWAP~\cite{zhang2018multi} & 43.0 & 57.8 & 61.9 & - \\
& DenseWAP-TD~\cite{zhang2020treedecoder} & 49.1 & 64.2 & 67.8 & - \\
& WS-WAP~\cite{truong2020improvement} & 53.65 & - & - & - \\
& Li \etal~\cite{li2020improving} & 56.59 & 69.07 & 75.25 & \textbf{78.60} \\
& Ding \etal~\cite{ding2021encoder} & 58.72 & - & - & - \\
& BTTR~\cite{zhao2021handwritten} & 53.96 & 66.02 & 70.28 & - \\
& BTTR (baseline) & 55.17 & 67.85 & 72.11 & 74.14 \\
\cmidrule{2-6}
& CoMER & \textbf{59.33} & \textbf{71.70} & \textbf{75.66} & 77.89 \\

\midrule

\multirow{8}{*}{CROHME 16}
& DenseWAP~\cite{zhang2018multi} & 40.1 & 54.3 & 57.8 & - \\
& DenseWAP-TD~\cite{zhang2020treedecoder} & 48.5 & 62.3 & 65.3 & - \\
& WS-WAP~\cite{truong2020improvement} & 51.96 & 64.34 & 70.10 & 72.97 \\
& Li \etal~\cite{li2020improving} & 54.58 & 69.31 & 73.76 & 76.02 \\
& Ding \etal~\cite{ding2021encoder} & 57.72 & 70.01 & 76.37 & 78.90 \\
& BTTR~\cite{zhao2021handwritten} & 52.31 & 63.90 & 68.61 & - \\
& BTTR (baseline) & 56.58 & 68.88 & 74.19 & 76.90 \\
\cmidrule{2-6}
& CoMER & \textbf{59.81} & \textbf{74.37} & \textbf{80.30} & \textbf{82.56} \\

\midrule

\multirow{6}{*}{CROHME 19}
& DenseWAP~\cite{zhang2018multi} & 41.7 & 55.5 & 59.3 & - \\
& DenseWAP-TD~\cite{zhang2020treedecoder} & 51.4 & 66.1 & 69.1 & - \\
& Ding \etal~\cite{ding2021encoder} & 61.38 & 75.15 & 80.23 & 82.65 \\
& BTTR~\cite{zhao2021handwritten} & 52.96 & 65.97 & 69.14 & - \\
& BTTR (baseline) & 59.55 & 72.23 & 76.06 & 78.40 \\
\cmidrule{2-6}
& CoMER & \textbf{62.97} & \textbf{77.40} & \textbf{81.40} & \textbf{83.07} \\

\bottomrule
\end{tabular}
\label{tb:sota}
\end{center}
\end{table}

\subsection{Performance at Different Lengths}
\begin{figure}[htbp]
	\centering
	\includegraphics[width=\textwidth]{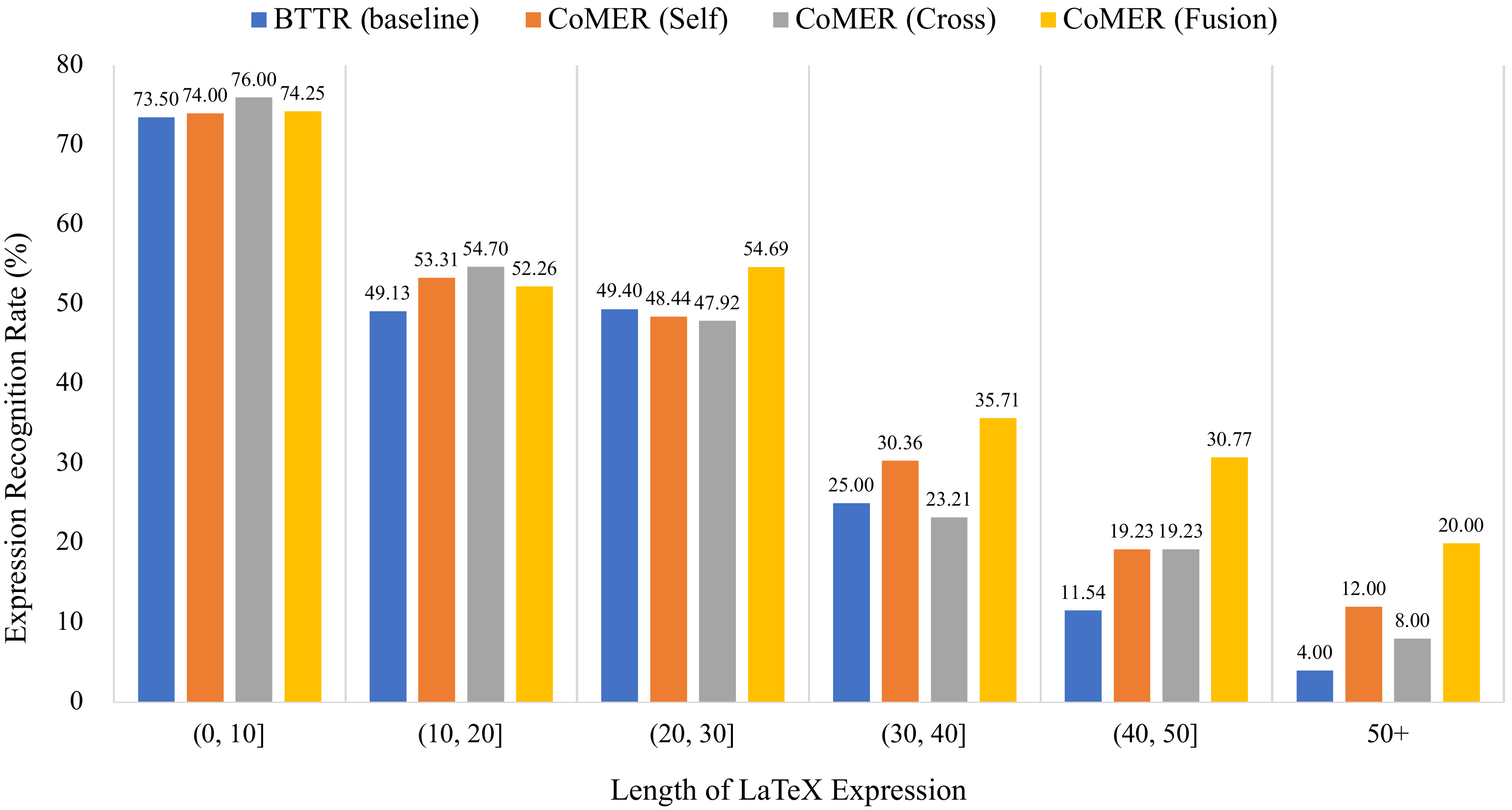}
	\caption{Recognition accuracy (in \%) for \LaTeX{} expressions of different lengths on the CROHME 2014 test set.}
	\label{fig:length}
\end{figure}

Intuitively, we assume that the recognition accuracy of long sequences is lower than that of short ones because of the lack of coverage problem~\cite{ZhaopengTu2016ModelingCF, zhang2017watch}. Thus, we consider the recognition accuracy of long sequences reflects the ability of a model to align the sequence and image. To verify that CoMER has better alignment and thus alleviates the lack of coverage problem, we calculate the recognition accuracy at different lengths on the CROHME 2014 test set, shown in Fig.~\ref{fig:length}.

By comparing ``BTTR (baseline)" with the CoMER family of models, we found that CoMER equipped with ARM has better performance when dealing with various lengths of sequences, especially with longer ones. The performance of ``CoMER (Fusion)" can reach even $\mathbf{5\times}$ than ``BTTR (baseline)" when recognizing sequences longer than 50. This experiment results show that the ARM and coverage mechanisms can improve the alignment quality and mitigates the lack of coverage problem.

Comparing the performance between self-coverage and cross-coverage, we find that cross-coverage performs better when parsing short sequences. In contrast, self-coverage is better at recognizing long sequences. We suppose this is because cross-coverage accumulates misalignments generated by the previous layer, causing it to incorrectly refine the attention weights in the current layer. In comparison, self-coverage performs alignment and refinement in each layer independently. ``CoMER (Fusion)" uses both self-coverage and cross-coverage to exploit diverse alignment information and far outperforms other models in recognizing sequences longer than 20.

\subsection{Refinement Term Visulization}
\begin{figure}[htbp]
	\centering
	\includegraphics[width=\textwidth]{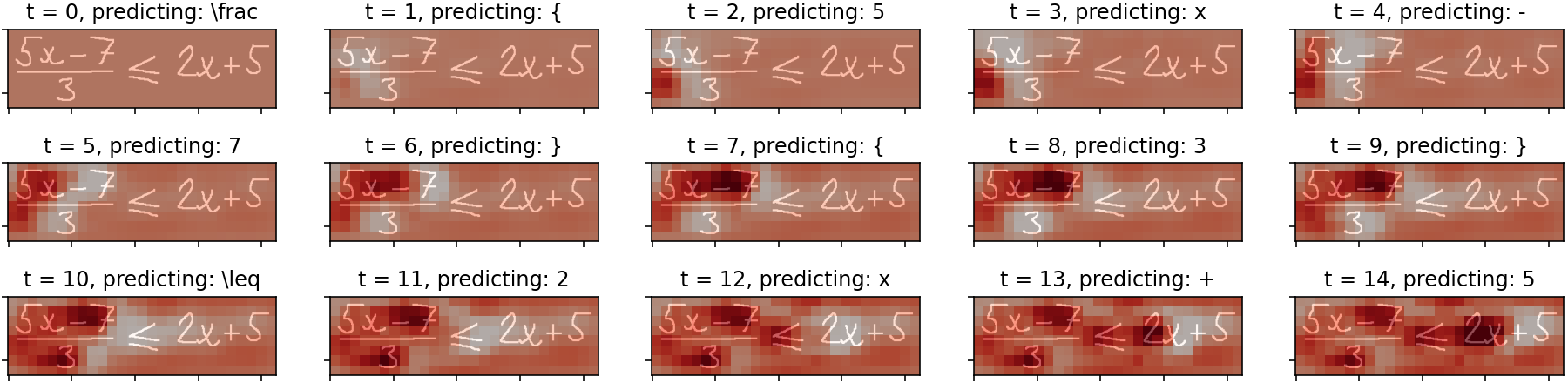}
	\caption{The refinement term $\mathbf{R}$ visualization in recognizing formula image. The darker the color, the larger the value.}
	\label{fig:refine-vis}
\end{figure}

As shown in Fig.~\ref{fig:refine-vis}, we visualize the refinement term $\mathbf R$ in the recognition process. We find that parsed regions are darker, which indicates ARM will suppress the attention weights in these parsed regions and encourage the model to focus on incoming unparsed regions. The visualization experiment shows that our proposed ARM can effectively alleviate the lack of coverage problem.

\section{Conclusion and Future Work}
In this paper, inspired by the coverage attention in RNN, we propose CoMER to introduce the coverage mechanism into the transformer decoder. We have the following four main contributions: \textbf{(1)} Our proposed CoMER alleviates the lack of coverage problem and significantly improves the recognition accuracy of long \LaTeX{} expressions. \textbf{(2)} We propose a novel Attention Refinement Module (ARM) that makes it possible to perform attention refinement in the transformer without harming its parallel computing nature. \textbf{(3)} We propose self-coverage, cross-coverage, and fusion-coverage to refine the attention weights using the past alignment information from the current and previous layers. \textbf{(4)} Experiments demonstrate the effectiveness of our proposed CoMER model. Specifically, we achieve new state-of-the-art performance on the CROHME 2014/2016/2019 test sets using a single CoMER model, reaching 59.33\%/59.81\%/62.97\% in ExpRate.

We believe that our proposed attention refinement framework not only works for handwritten mathematical expression recognition. Our ARM can help refine the attention weights and improve the alignment quality for all tasks that require dynamic alignment. To this end, we intend to extend ARM in the transformer as a general framework for solving various vision and language tasks in the future work (e.g., machine translation, text summarization, image captioning).

\subsubsection*{Acknowledgements.}
This work is supported by the projects of National Key R\&D Program of China (2019YFB1406303) and National Nature Science Foundation of China (No. 61876003), which is also a research achievement of Key Laboratory of Science, Technology and Standard in Press Industry (Key Laboratory of Intelligent Press Media Technology).

\bibliographystyle{splncs04}
\bibliography{main}
\end{document}